\documentclass[twoside,twocolumn]{article}

\usepackage{blindtext} 

\usepackage[sc]{mathpazo} 
\usepackage[T1]{fontenc} 
\usepackage{microtype} 

\usepackage[english]{babel} 

\usepackage[hmarginratio=1:1,top=32mm,columnsep=20pt]{geometry} 
\usepackage[hang, small,labelfont=bf,up,textfont=it,up]{caption} 
\usepackage{booktabs} 

\usepackage{lettrine} 

\usepackage{enumitem} 
\setlist[itemize]{noitemsep} 

\usepackage{abstract} 

\usepackage{titlesec} 
\renewcommand\thesection{\Roman{section}} 
\renewcommand\thesubsection{\roman{subsection}} 
\titleformat{\section}[block]{\large\scshape\centering}{\thesection.}{1em}{} 
\titleformat{\subsection}[block]{\large}{\thesubsection.}{1em}{} 

\usepackage{fancyhdr} 
\usepackage{titling} 

\usepackage{hyperref} 

\usepackage{graphicx}
\usepackage{url}

\pretitle{\begin{center}\Huge\bfseries} 
\posttitle{\end{center}} 
\title{How Intelligent is your Intelligent Robot?} 
\author{%
\textsc{Alan F. T. Winfield}
\\[1ex] 
\normalsize Bristol Robotics Lab \\ 
\normalsize \href{mailto:alan.winfield@uwe.ac.uk}{alan.winfield@uwe.ac.uk} 
}
\date{} 
\renewcommand{\maketitlehookd}{%

\begin{abstract}
\noindent 
How intelligent is robot A compared with robot B? And how intelligent are robots A and B compared with animals (or plants) X and Y? These are both interesting and deeply challenging questions. In this paper we address the question "how intelligent is your intelligent robot?" by proposing that embodied intelligence emerges from the interaction and integration of four different and distinct kinds of intelligence. We then suggest a simple diagrammatic representation on which these kinds of intelligence are shown as four axes in a star diagram. A crude qualitative comparison of the intelligence graphs of animals and robots both exposes and helps to explain the chronic intelligence deficit of intelligent robots.  Finally we examine the options for determining numerical values for the four kinds of intelligence in an effort to move toward a quantifiable intelligence vector.

\end{abstract}
}

\begin{document}

\maketitle


\section{Introduction}

\lettrine[nindent=0em,lines=3]{B}enchmarking the performance of robots is notoriously difficult, even when comparing similar robots undertaking the same task under controlled conditions \cite{Fontana2014}. But if we want to ask a different question: "how intelligent is any given robot?" we face a much bigger challenge, especially if we want to position robot intelligence within the spectrum of animal (or more generally) natural intelligence. The first challenge is that there is no satisfactory pan-species definition of intelligence, i.e. one that we could apply to all animals as well as robots. The second is that intelligence is not one thing that animals or robots have more or less of.

This paper will attempt to make headway in addressing the problem of assessing robot intelligence in several ways. First, we propose four categories of intelligence: morphological intelligence, swarm intelligence, individual intelligence and social intelligence. We then suggest a simple diagrammatic representation on which these kinds of intelligence are shown as four axes in a star diagram. Guessing at relative values for certain animals and robots and superimposing these immediately and surprisingly reveals why present day robot intelligence falls so far behind animal intelligence. The paper then examines the options for determining numerical values for the four kinds of intelligence, in an effort to move toward a quantifiable intelligence vector.


\section{Kinds of Intelligence}

How do we define intelligence? Is is as simple as `doing the right thing at the right time' or is learning (or adaptation) so fundamental that any definition must encompass learning? Legg and Hutter \cite{LeggHutter2007} collected 70 different definitions for intelligence, which they distilled into ``Intelligence measures an agent's ability to achieve goals in a wide range of environments"; they note that the ability to learn or adapt is implicit in this definition.

In this paper we sidestep the thorny problem of coming up with a singular definition of intelligence by proposing that embodied intelligence \textit{inter alia} emerges from the interaction and integration of four different and distinct kinds of intelligence. These are morphological intelligence and swarm intelligence, which fall upon one axis which we might label body/bodies, and individual and social intelligence that we might label adaptation. These four kinds of intelligence are defined as follows.

\subsection{Morphological Intelligence}

This is the kind of intelligence that a physical body confers to its owner. The idea that a body has some inherent intelligence may seem odd, but is closely related to the notion of morphological computation; which has been defined as ``a term which captures conceptually the observation that biological systems take advantage of their morphology to conduct computations needed for a successful interaction with their environments" \cite{Hauser2013}. Following Pfeifer and Iida \cite{PfeiferIida2005} let us define morphological Intelligence as the \textit{physical} behaviour that emerges from the interaction of body, its control system and the environment. The best way to illustrate morphological intelligence is with simple robots that are notably unintelligent. David Buckley's minimal walking robot TechFoot, for example, uses only 5 servomotors but achieves bipedal locomotion by exploiting both natural oscillations and energy storage in the leg and hip structure to achieve dynamically stable walking \cite{Buckley2000}. For another example consider the Solarbot, a simple Braitenberg machine \cite{Braitenburg1986} with just two solar panels and two motors, in which the left hand solar sensor drives the right hand motor, and vice versa. Despite having no computational control, Solarbot demonstrates surprisingly rich behaviours \cite[p. 43]{Winfield2012}. A third and even more dramatic example is the Jeager-Lipson coffee balloon gripper \cite{Brownetal2010}. A gripper based on a balloon filled with coffee grounds, which exploits the phase transition from fluid to solid when the air is evacuated from the gripper. A wonderful example of soft robotics, the Jaeger-Lipson gripper shows huge versatility in its ability to reliably grip a wide range of objects yet requires none of the morphological or control complexity of robot grippers modelled on the human hand.

\subsection{Swarm Intelligence}

Swarm Intelligence describes the collective, self-organised behaviour we observe in animals that swarm, shoal, flock or herd, or - more dramatically - build complex nest structures such as ants, bees, or termite mounds. Swarm intelligence is an emergent property of the collective that results from the local interactions of the individuals with each other and with their environment \cite{DorigoBirattari2007}. The individuals of the swarm act autonomously on the basis of local sensing and local communications only, and importantly there is no command and control hierarchy in a swarm - no `brain' ant in an ant colony - instead control is completely distributed and decentralised.

Swarm robotics is an effort to create swarms of robots in which the desired collective behaviours emerge from the local interactions between individual robots and their environment \cite{SahinWinfield2008}. Several collective behaviours have been successfully demonstrated in the laboratory, including flocking, foraging \cite{Liu2010b}, collective transport \cite{Gross2009} and collective construction \cite{Liu2010},\cite{Werfel2014}. Indeed  foraging is a canonical problem in swarm robotics, since foraging is a metaphor for many potential real world applications of robot swarms \cite{Winfield2009}.

\subsection{Individual Intelligence}

Individual intelligence is defined here as the ability to both respond (instinctively) to stimuli and, optionally, learn new -- or adapt existing -- behaviours through, typically, a process of trial and error. The instinctive component of individual intelligence is hard wired; in plants and animals it is evolved or, in robots, either hand designed or artificially evolved. If a learned component is present the actual learning mechanism is not important here, except that it must be the individual that learns in its own lifetime, without the help of another individual.

Dennett proposes an elegant conceptual model of adaptation in his Tower of Generate and Test \cite{Dennett1995}. Dennett argues that adaptation requires actions (in response to some stimulus) to be generated and tested by an animal, and classifies animals according to the mechanism employed. Dennett's tower is set of conceptual creatures each one of which is successively more capable of reacting to (and hence surviving in) the world through having more sophisticated strategies for `generating and testing' hypotheses about how to react. On the first floor of Dennett's tower are Darwinian creatures, which have only natural selection as the generate and test mechanism, so variation and selection is the only way that Darwinian creatures can adapt -- individuals cannot. On the second floor are Skinnerian creatures, which can learn but only by literally generating and testing all different possible actions then reinforcing the successful behaviour. 
On the third floor of the tower are Popperian creatures which have the  additional ability to internally model the possible actions so that some (the bad ones) are rejected before they are tried out for real. Like the Tower of Hanoi each successive storey is smaller (a sub-set) of the storey below, thus only a sub-set of Darwinian creatures are Skinnerian and so on. In the scheme of this paper we regard Dennett's Darwinian, Skinnerian and Popperians creatures as all exhibiting individual intelligence.

There are many robots with individual intelligence of different types; examples include reactive behaviours such as obstacle avoidance which has been designed by hand \cite{Boren1989}, or artificially evolved \cite{Barate2009}. There are also many examples of robots capable of individual learning: one important class of learning algorithm allows a robot to simultaneously localise itself within its environment while building a map of that environment (SLAM) \cite{Thrun2008}, another is reinforcement learning (RL), for a survey see \cite{Kober2013}.

\subsection{Social Intelligence}

Social intelligence is the kind of intelligence that allows animals or robots to learn from each other. This might be through imitation or instruction. In imitation a new behaviour is acquired by the social learner observing another's behaviour then transforming those observations into corresponding actions and responses. With or without the active cooperation of the demonstrator, imitation is a complex process which requires cognitive mechanisms to solve the so called correspondence problem \cite{Nehaniv2002}; how best to match observed sense data with corresponding motor actions. Social learning through instruction is even more complex since it requires a teacher to explain a behaviour using shared symbols, 
and the learner to properly interpret and assimilate those instructions.

Returning to Dennett's tower of generate and test, on the fourth floor we find Gregorian creatures. These are tool makers including - importantly - mind tools like language, which means that individuals no longer have to generate and test all possible hypotheses since others have done so already and can pass on that knowledge. Gregorian creatures 
are tool makers, of both physical tools (like scissors) and mind-tools (like language and mathematics), and Dennett suggests that these tools are `intelligence amplifiers'. Certainly they give Gregorian creatures a significant advantage over Popperians, because they have the benefit of the shared experience of others, expressed either through using the tools they have made or refined or, more directly, through their knowledge or instructions as spoken or written.

The study of imitation and social learning in robots, humans and animals has received cross-disciplinary attention 
\cite{Nehaniv2007}. Not surprisingly much attention has been given to the problem of humanoid robots imitating humans, since this presents a way of programming a robot by demonstration (PbD) rather than coding \cite{Mataric2000}, \cite{Billard2013}. There has been less work in robot-robot imitation. The earliest is perhaps the work of Hayes and Demiris which describes an approach with one (pre-programmed) robot finding its way through a maze and another following it and observing its actions (turns). The following (learner) robot then associates each observed action with its own -- time delayed -- perception of the environment and hence learns how to navigate the maze, by imitation \cite{Hayes1994}. More recent work demonstrates  embodied behavioral evolution in a group of robots through repeated cycles of imperfect robot-robot imitation, in which the process of noisy imitation introduces variations which, in turn, introduce behavioural novelty \cite{Winfield2011}.


\section{A map of intelligence}

\begin{figure}[h!]
	\begin{center}
		\includegraphics[width=7.5cm]{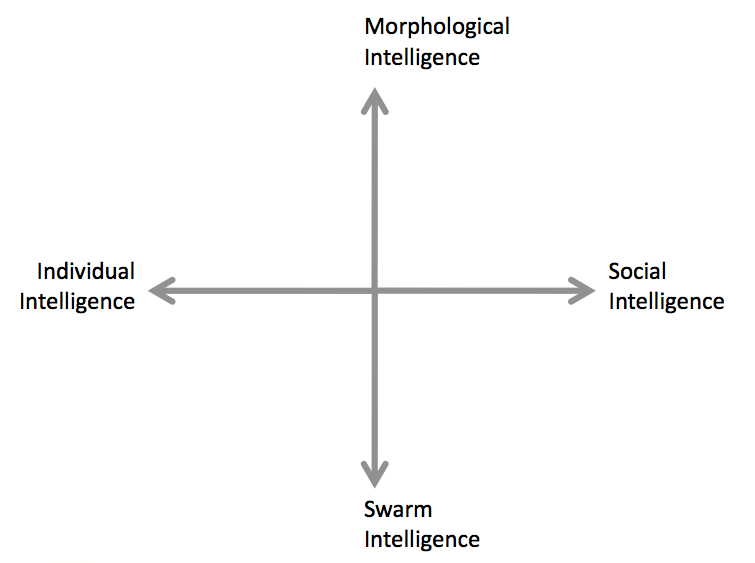}
	\end{center}
 	\textbf{\refstepcounter{figure}\label{fig:01} Figure 		\arabic{figure}.}{ The axes of intelligence}
\end{figure}

Let us now put these four kinds of intelligence together as two axes, in Fig. \ref{fig:01}. We place morphological and swarm intelligence on the vertical body/bodies axis and individual and social intelligence on the horizontal adaptation axis.

\begin{figure}[h!]
	\begin{center}
		\includegraphics[width=7.5cm]{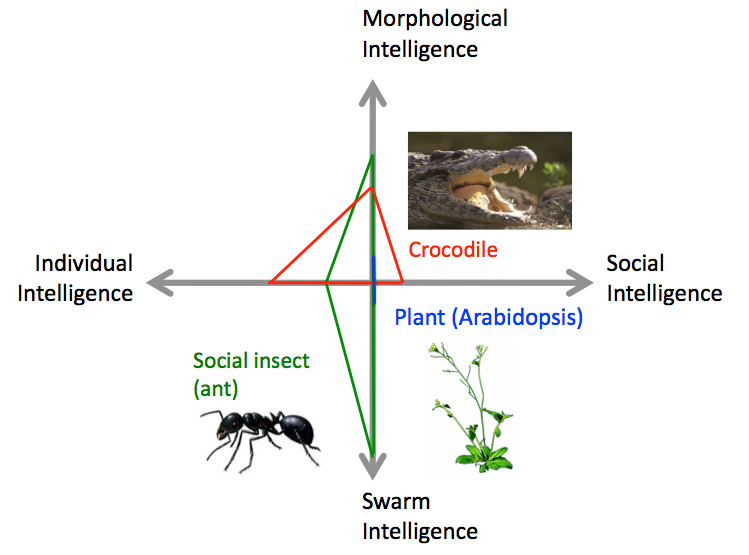}
	\end{center}
 	\textbf{\refstepcounter{figure}\label{fig:02} Figure 		\arabic{figure}.}{ Comparing animal and plant intelligence}
\end{figure}

Fig. \ref{fig:02} attempts a comparison of animal and plant intelligence. We suppose a crocodile to exhibit morphological and individual intelligence, and a limited amount of social intelligence, but no swarm intelligence - as far as we know crocodiles do not shoal or herd. In contrast ants have no social intelligence (as far as we know), but considerable swarm intelligence, and somewhat more morphological intelligence than a crocodile, since its body is more complex; ants also exhibit some individual intelligence, including simple learning. Plants certainly exhibit morphological intelligence 
as well as some elements of swarm intelligence 
\cite{Cham2012}. 

\begin{figure}[h!]
	\begin{center}
	\includegraphics[width=7.5cm]{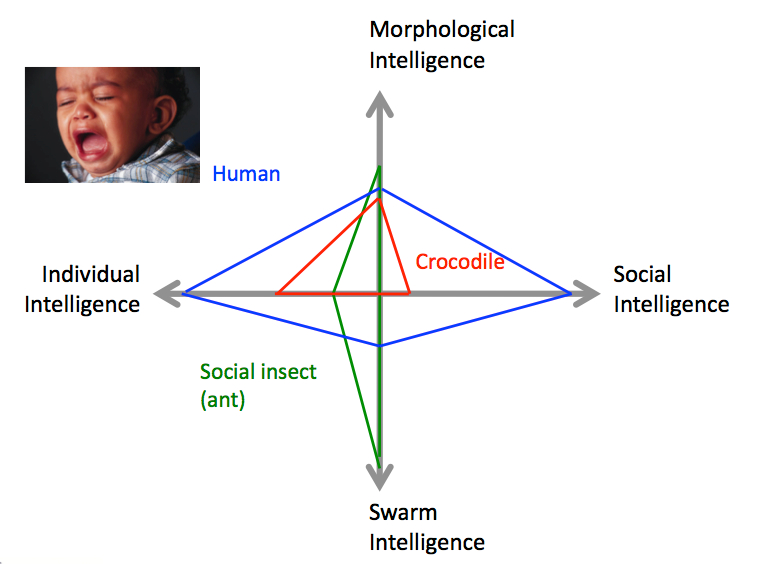}
	\end{center}
 \textbf{\refstepcounter{figure}\label{fig:03} Figure 		\arabic{figure}.}{ Comparing human and animal intelligence}
\end{figure}

If we now add humans to this comparison, in Fig. \ref{fig:03}, we see that humans exhibit huge levels of individual and social intelligence, presumably more than any other known animal. But a human's morphological intelligence is arguably somewhat lower than an ant's, and while humans do exhibit a degree of swarm intelligence -- as crowd behaviour -- this is at a much lower level of significance than social insects.

Of course the intelligence maps represented by the star diagrams of Figs \ref{fig:02} and \ref{fig:03} are crude guesses; the simple fact is that we cannot assign values to the levels of each of our four categories of intelligence, any better than we can guess the IQ of a crocodile. Nor can we have any confidence in the relative values shown here (does a crocodile really have half the individual intelligence of a human?): an observation that leads to the suggestion that these axes might be better constructed as logarithmic scales.

These star diagrams do, however, have explanatory value. Recalling our working definition of intelligence as an emergent property of the interaction and integration of four different and distinct kinds of intelligence, the star diagrams serve to illuminate clear deficits as well as relative strengths in the kinds of intelligence apparently demonstrated by living organisms. 

\begin{figure}[h!]
	\begin{center}
	\includegraphics[width=8cm]{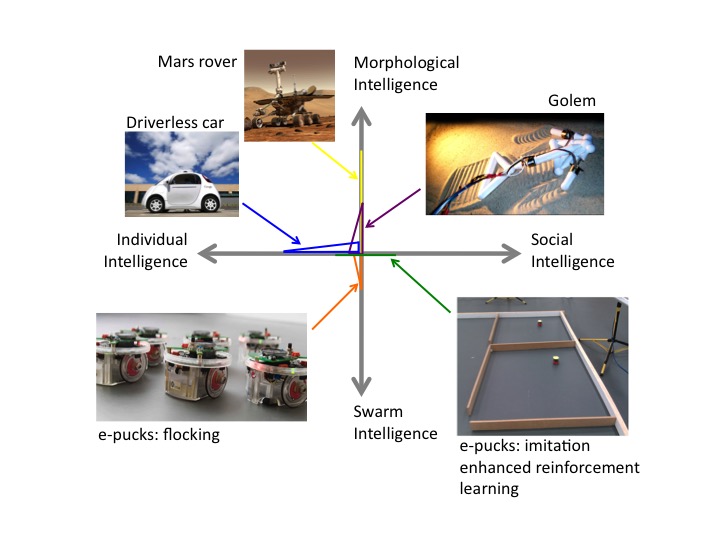}
	\end{center}
 \textbf{\refstepcounter{figure}\label{fig:04} Figure 		\arabic{figure}.}{ How do robots measure up?}
\end{figure}

Now let us turn to the question of how robots measure up within this scheme. Fig. \ref{fig:04} attempts a similar estimation of the relative levels of our four kinds of intelligence for a small number of exemplar robots. Of course we can be more confident of the kinds of intelligence expressed by these robots since they are designed (or, in one case, evolved) artefacts, although the relative levels of intelligence shown in fig. \ref{fig:04} remain guesses.

Starting on the bottom left is a swarm of e-puck robots exhibiting perhaps the simplest of swarm behaviours: flocking. With only 2 wheels these robots are morphologically very simple and therefore merit no score for morphological intelligence, but since each robot is individually capable of simple obstacle avoidance they do show a very low level of individual intelligence -- hence the orange graph in fig. \ref{fig:04}. The illustration on the bottom right shows one of the few examples of robots that exhibit both individual (reinforcement) learning -- in this case learning how to reach a goal position in the arena -- and social learning -- periodically each robot will leave its arena and observe the other to imitate its behaviour; we call this imitation-enhanced reinforcement learning \cite{Erbas2014}. The intelligence graph for the e-puck robots executing these behaviours is shown in green and extends across the individual and social intelligence axes.

In the top right of fig. \ref{fig:04} we see the Golem robot, notable because both its body shape and (artificial neural network) controller are artificially evolved for locomotion with linear actuators which exploit frictional forces \cite{Lipson2000}. Its intelligence graph is shown in purple with a reasonable score for morphological intelligence and a small score for individual intelligence (although whether this robot merits any score at all for individual intelligence is moot - since the ANN which coordinates its actuators is an integral and necessary part of its morphological intelligence). In the top left of fig. \ref{fig:04} is a Mars rover, included here because it exhibits a high level of morphological intelligence (shown in yellow) thanks to the ingenious design of its rocker-bogey mounted wheels. These give the rover the ability to cross rough terrain, including modestly sized rocks, without explicit control, since the arrangement of wheels and bogeys simply deforms to accommodate those obstacles.

Finally, as a topical example, we consider the question: how intelligent is a driverless car? Morphologically the driverless car shown in fig. \ref{fig:04} is very simple indeed -- certainly much simpler than the Mars rover and Golem robot -- with four wheels and standard front-wheel Ackermann steering. However, in the intelligence graph (shown in blue) we acknowledge that the tyres fitted to the driverless cars' wheels, together with its suspension, afford the car some ability to deal with uneven road surfaces without explicit control. The driverless car's autopilot provides it with a considerable degree of individual intelligence in managing the  complexity of safe autonomous driving -- even in favourable traffic and weather conditions; this almost certainly makes the driverless car one of the most (individually) intelligent robots to date.

\section{Why Intelligent Robots are not very}

Our intuition leads us to the view that, in general, the intelligence of intelligent robots falls far short of that of most animals. Fig. \ref{fig:04} exposes at least one of the reasons for this intelligence gap: none of the intelligence graphs for the exemplar robots score on more than two axes, whereas all of the exemplar animals in fig. \ref{fig:03} score on at least three axes. Of course neither figs. \ref{fig:03} or \ref{fig:04} represent a comprehensive survey, but nevertheless we are drawn to the provisional conclusion that robots typically exhibit some \textit{limited} morphological intelligence plus some \textit{limited} form of one of the other kinds of intelligence (the example of e-puck robots demonstrating imitation-enhanced reinforcement learning \cite{Erbas2014} is a rare exception). Why limited? It only needs a moment's reflection to appreciate that, firstly, the level of morphological sophistication demonstrated in robots falls far below that of most organisms and, secondly, that the scope and sophistication of the singular kinds of intelligence exhibited by robots is very narrow. 

In order to make a living all animals must and do successfully forage or hunt for food, evade predators and find a mate, while some build nests, nurture young, construct social hierarchies and/or seasonally migrate; some make and use tools. 
No robot demonstrates even a small fraction of these basic survival skills.
The fact that the levels of intelligence in the graphs of fig. \ref{fig:04} are so chronically low suggests that we should imagine magnifying the region of fig. \ref{fig:03} close to the origin in its centre in order to make a meaningful comparison between the two figures.

Let us now speculate on how we might construct more accurate quantitative assessments of the four kinds of intelligence in order to construct an intelligence vector for any given intelligent agent. At first this might appear to be a hopeless endeavour (and for all but the simplest animals it may well be). But one thing all animals and autonomous robots have in common is \textit{action-selection} \cite{Seth2013}: all autonomous agents must, from time to time, choose their next action -- normally in response to some sensory stimulus -- from a pool of next possible actions. This observation suggests a crude metric: we could simply count the number of next possible actions. For robots at least obtaining such a measure should be feasible. This suggestion immediately raises a number of difficult questions including, first, how do we define a unit of action: is it low-level such as `actuate left motor', mid-level like `avoid obstacle' or the higher-level `navigate safely to battery recharging station'? Second, how would such actions be identified for the category of morphological intelligence, and third, how do we define discrete emergent swarm behaviours (flocking, foraging, etc)? 

\section{Conclusions} \label{Conclusions}
In this paper we have proposed a new approach to addressing the question: how intelligent are intelligent robots, by proposing that intelligence emerges from the interaction and integration of four kinds of intelligence: morphological, swarm, individual and social. By placing these four kinds of intelligence in a map (star diagram) we can guess a set of intelligence graphs for different animals (and plants) and exemplar robots. These very approximate graphs expose the chronic intelligence deficit of (intelligent) robots when compared with animals, revealing in particular the fact that most robots demonstrate only two kinds of intelligence (morphological plus one other) in stark contrast with animals which typically exhibit high levels of both morphological intelligence and two or three other kinds of intelligence.

Why is this? One is the obvious conclusion that the development of intelligent robots is an immature science: an early work-in-progress. Another is that labs typically focus their research on one kind of intelligence; few if any are actively pursuing the hard task of integrating all kinds of intelligence in a single robot -- something Dennett \cite{Dennett1978} and Brooks \cite{Brooks1989} describe as ``building the whole iguana''.

\section*{Acknowledgements}
Development of the ideas in this paper have benefited from discussion with many colleagues, including especially Susan Blackmore, Jose Halloy and Roger Moore.


\end{document}